\documentclass[letterpaper, 10 pt, conference]{ieeeconf}  % Comment this line out if you need a4paper

\IEEEoverridecommandlockouts                              % This command is only needed if 
\usepackage{bm}
\usepackage{url}
\usepackage{graphicx}
\usepackage{epstopdf}
\usepackage{lipsum}
\usepackage{pifont}
\usepackage{amsmath}
\usepackage{amssymb}
\usepackage{nccmath}
\usepackage{comment}
\usepackage{balance}
\usepackage{multirow}
\usepackage{textcomp}
\usepackage{multirow}
\usepackage{subcaption}
\usepackage{booktabs}
\usepackage{mathtools}
\usepackage{algorithm}
\usepackage{dblfloatfix}
\usepackage{xcolor}
\usepackage[noend]{algpseudocode}

\title{\LARGE \bf
Range-based 6-DoF Monte Carlo SLAM \\
with Gradient-guided Particle Filter on GPU
}

\author{Takumi Nakao$^{1}$, Kenji Koide$^{2}$, Aoki Takanose$^{2}$, Shuji Oishi$^{2}$, Masashi Yokozuka$^{2}$, and Hisashi Date$^{3}$
\thanks{*This work was supported in part by JSPS KAKENHI Grant Number 23K03896 and the project JPNP14004 subsidized by the New Energy and Industrial Technology Development Organization (NEDO).}% <-this % stops a space
\thanks{$^{1}$Takumi Nakao is with the Graduate School of Science and Technology, University of Tsukuba, Tsukuba, Ibaraki, Japan, {\tt\small nakao-t@roboken.iit.tsukuba.ac.jp}}%
\thanks{$^{2}$Kenji Koide, Aoki Takanose, Shuji Oishi, and Masashi Yokozuka are with the National Institute of Advanced Industrial Science and Technology, Tsukuba, Ibaraki, Japan}%
\thanks{$^{3}$Hisashi Date is with the Institute of Systems and Information Engineering, University of Tsukuba, Tsukuba, Ibaraki, Japan}%
}

\begin{document}

\maketitle
\thispagestyle{empty}
\pagestyle{empty}

\setlength\floatsep{8pt}
\setlength\textfloatsep{8pt}

%%%%%%%%%%%%%%%%%%%%%%%%%%%%%%%%%%%%%%%%%%%%%%%%%%%%%%%%%%%%%%%%%%%%%%%%%%%%%%%%
\begin{abstract}

This paper presents range-based 6-DoF Monte Carlo SLAM with a gradient-guided particle update strategy. While non-parametric state estimation methods, such as particle filters, are robust in situations with high ambiguity, they are known to be unsuitable for high-dimensional problems due to the curse of dimensionality. To address this issue, we propose a particle update strategy that improves the sampling efficiency by using the gradient information of the likelihood function to guide particles toward its mode. Additionally, we introduce a keyframe-based map representation that represents the global map as a set of past frames (i.e., keyframes) to mitigate memory consumption. The keyframe poses for each particle are corrected using a simple loop closure method to maintain trajectory consistency. The combination of gradient information and keyframe-based map representation significantly enhances sampling efficiency and reduces memory usage compared to traditional RBPF approaches. To process a large number of particles (e.g., 100,000 particles) in real-time, the proposed framework is designed to fully exploit GPU parallel processing. Experimental results demonstrate that the proposed method exhibits extreme robustness to state ambiguity and can even deal with kidnapping situations, such as when the sensor moves to different floors via an elevator, with minimal heuristics.

\end{abstract}

%%%%%%%%%%%%%%%%%%%%%%%%%%%%%%%%%%%%%%%%%%%%%%%%%%%%%%%%%%%%%%%%%%%%%%%%%%%%%%%%
\section{INTRODUCTION}

Simultaneous localization and mapping (SLAM) is an essential function for autonomous systems such as service robots and autonomous vehicles. Most existing 3D SLAM algorithms are based on Kalman filters or factor graph-based state estimation~\cite{Bai2022, Grisetti2010}. These methods are highly efficient since MAP estimation can be simplified to solve basic linear algebra problems by locally approximating the nonlinear likelihood function as a Gaussian distribution~\cite{probrobo, dellaert2017factor}. However, their accuracy and robustness decline when the local Gaussian assumption becomes invalid due to inaccurate initial guesses or ambiguous observations. As a result, these approaches struggle in scenarios with high ambiguity, such as long-term degeneration of point clouds or loop closing with multiple uncertain candidates.

The particle filter is a non-parametric state estimation method that represents the state distribution with a finite set of samples (i.e., particles). Due to the flexibility of its sampling-based state representation, the particle filter is robust against state ambiguity in lower-dimensional problems (e.g., 2D SLAM~\cite{grisetti2007improved}).

\begin{figure}[tb]
 \centering
 \includegraphics[width=\linewidth]{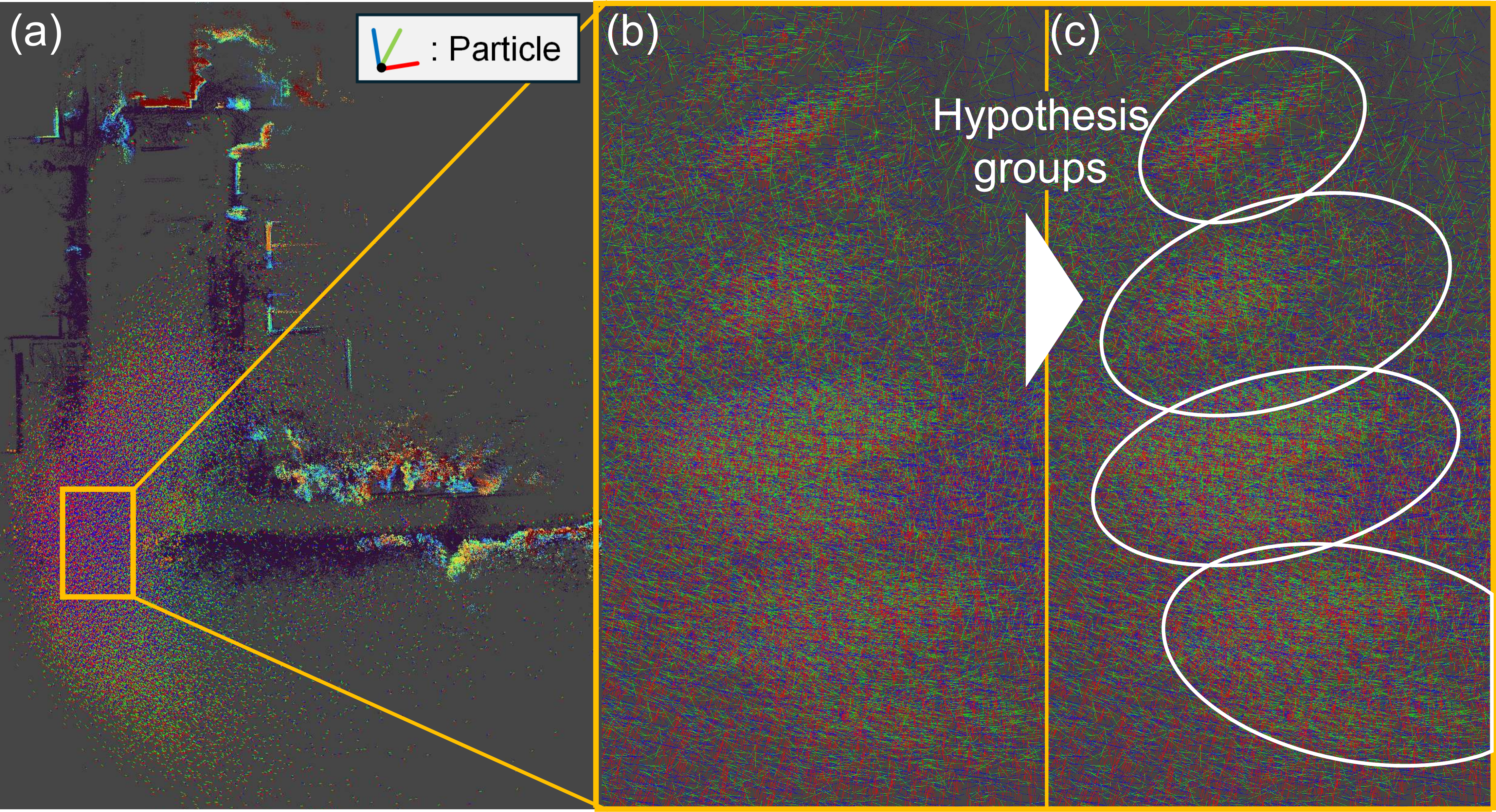}
 \caption{Estimated point cloud map and a set of 100,000 particles processed in real-time on a GPU (a). The very large number of particles enables flexible representation of multi-modal state distributions in highly ambiguous situations (e.g., in a forest-like environment) (b) (c).}
 \label{fig:thumbnail}
\end{figure}

Rao-Blackwellized Particle Filter (RBPF) SLAM is an efficient application of a particle filter to the SLAM problem, due to the factorization of the map from the posterior distribution. However, applying this approach to 3D SLAM problems has two major challenges. The first challenge is high computational complexity. As the state space's dimensionality and the environment's ambiguity increase, the required number of particles becomes large. This reason is that the number of samples needed to cover a unit space grows exponentially with dimensionality (i.e., the curse of dimensionality). Traditional Monte Carlo approaches suffer from a significant computational burden in high-dimensional problems. The second challenge is memory limitations due to map representation. Traditional RBPF methods represent a map as an occupancy grid, which significantly increases the memory footprint in 3D space. Furthermore, as each particle in RBPF holds a distinct trajectory and map, the total memory usage becomes intractable as the number of particles increases.

To improve particle sampling efficiency, several recent works have introduced particle update strategies based on variational inference in particle filters. These strategies guide particle states toward the mode of the likelihood function using gradient information while preserving sample diversity through neighbor particle information~\cite{Maken_2022,resampling_free_icra2024,Koide_2024mega}. These new particle filter algorithms have demonstrated a significant improvement in sampling efficiency, making them applicable to high-dimensional state estimation problems.

In this paper, we present a 6-DoF Monte Carlo SLAM algorithm for 3D environments, incorporating a gradient-guided particle filter with GPU acceleration and a keyframe-based compact map representation. There are three key points that distinguish the proposed method from traditional RBPF SLAM. First, we introduce a keyframe-based map representation that reduces memory usage, as each particle only needs to store keyframe poses and can share keyframe point clouds across all particles. Second, inspired by recent variational inference-based particle filters~\cite{Maken_2022, resampling_free_icra2024, Koide_2024mega}, we propose a particle filter that uses the gradient of the likelihood function to adjust the sensor pose estimate for each particle. This gradient information allows particles to be moved directly toward the mode of the likelihood, significantly improving sampling efficiency. Furthermore, unlike traditional RBPF, we correct the estimates of past sensor poses (i.e., keyframe poses) based on current pose updates to maintain trajectory consistency. Third, the entire system is designed to fully exploit the GPU, enabling real-time processing of 100,000 particles.

To showcase the ability to handle highly ambiguous situations, we conducted experiments in an outdoor forest-like environment with repeated rows of trees, which poses a challenge for loop closure. The experimental results showed that the proposed method can robustly correct trajectory errors even in environments with multiple uncertain loop candidates. We also tested the system by moving the sensor from one floor to another via an elevator. This is an extremely challenging scenario, as the floors have similar layouts, introducing significant ambiguity. To our knowledge, existing methods cannot handle such situations. Our experimental results demonstrated that the proposed method can accurately represent extreme ambiguity in localization and ultimately find the correct mapping solution once distinct features are observed later on.

The contribution of this work is three-fold:
\begin{itemize}
  \item To reduce memory consumption in 3D maps, we propose a keyframe-based map representation. The global map is represented as a union of keyframe point clouds, and each particle holds keyframe poses in addition to the current sensor pose as the estimation state. This approach significantly reduces the total memory footprint by sharing point cloud data across all particles.
  \item Inspired by recent variational inference-based particle filters~\cite{Maken_2022,Koide_2024mega}, we propose a method for updating particle states using the gradient of the likelihood function. This approach efficiently optimizes the state for each particle, including the current sensor pose and keyframe poses, compared to the traditional weighting-and-resampling approach, enabling 6-DoF mapping.
  \item To process a large number of particles (100,000 particles) in real-time, we designed the proposed system to fully leverage GPU parallel processing.
\end{itemize}

\section{RELATED WORK}
\begin{figure*}[tb]
 \centering
 \includegraphics[width=0.90\linewidth]{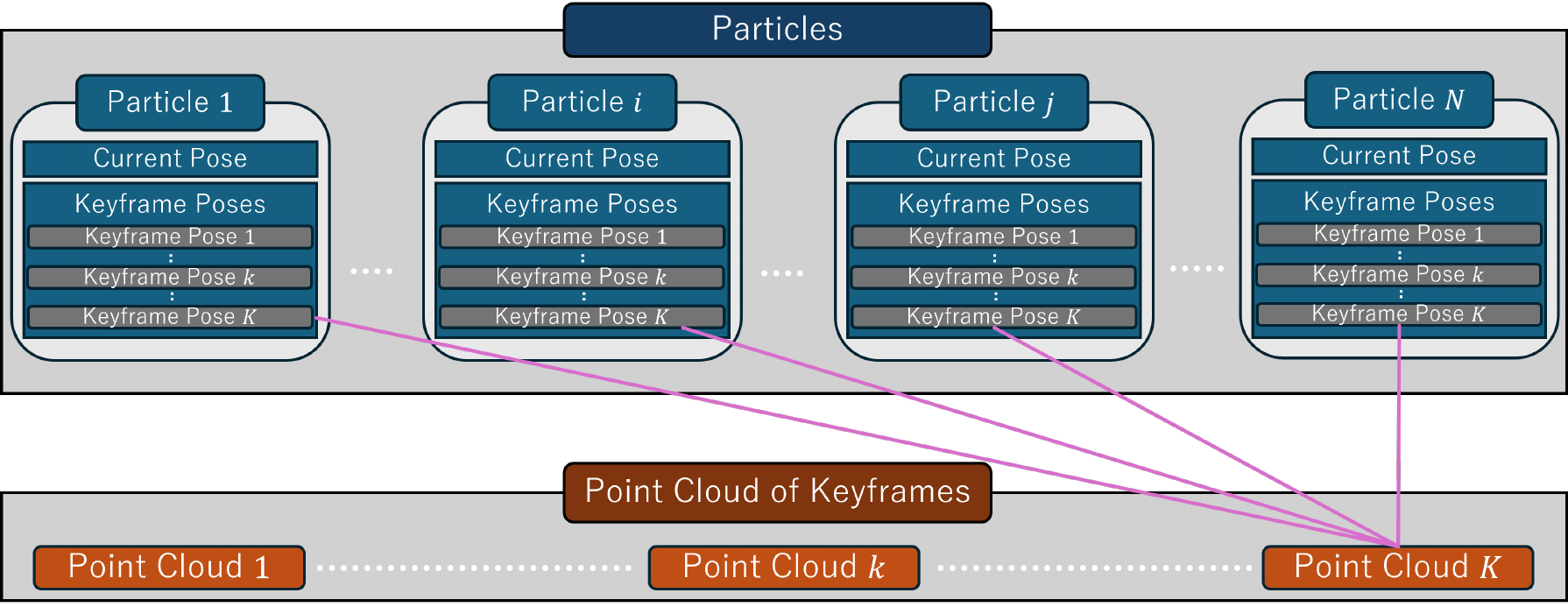}
 \caption{Data structure for particles and keyframes in the proposed system. Each of $N$ particles represents a hypothesis of the current pose and the keyframe poses. Separately from the particles, the point clouds of the keyframes are maintained in a global memory. The map estimate of each particle is given as a union of keyframe point clouds transformed with the estimate of keyframe poses.}
 \label{fig:system}
\end{figure*}
\subsection{Parametric State Estimation for SLAM}

Most of the existing 3D SLAM algorithms are based on state estimation with local Gaussian assumptions, such as Kalman filters and factor graph optimization. The Kalman filter and its non-linear extensions (extended Kalman filter~\cite{terejanu2008extended} and unscented Kalman filter~\cite{wan2000unscented}) are the most popular state estimation filters that represent the current state and observations as Gaussian distributions and propagate uncertainties through recursive Bayesian estimation. Since the Kalman filter only holds and estimates the current state and marginalizes out past states every time new observations arrive, it tends to accumulate linearization errors quickly~\cite{strasdat2012visual}. However, due to its efficiency, the Kalman filter has been preferred in real-time applications such as LiDAR-IMU odometry estimation~\cite{Xu2022,Chen_2023}. The factor graph is a powerful framework for Bayesian estimation based on a graph representation~\cite{dellaert2017factor}. This approach makes it possible to efficiently solve MAP estimation by reducing the optimization problem, which is represented as a graph, to a sparse linear algebra problem. While this approach also models state distributions and observations locally as Gaussian distributions, it is more accurate compared to the filtering approach as it re-linearizes sensor states at every optimization step and can handle many states due to the sparse representation. Since it is efficient and accurate, factor graph optimization has been widely used for many state estimation problems including visual and LiDAR odometry estimation~\cite{Wisth2023} and global trajectory optimization with loop closure~\cite{Shan2020,Nguyen_2023}. 

While these methods have been dominating the state estimation problems due to their efficiency, the local Gaussian assumption poses challenges to handling highly ambiguous situations. The Gaussian assumption can easily be violated when, for example, the initial guess is not accurate, or multiple plausible data associations exist. In the case of SLAM, such ambiguous situations commonly happen, for example, when the sensor moves very quickly or similar environments are repeated multiple times. The approaches based on the local Gaussian assumption inherently suffer from such situations.

Although several works have tackled multi-modal distributions by extending these approaches, such as the Gaussian sum Kalman filter~\cite{kottakki2014improved}, Multi-hypothesis iSAM2~\cite{fourie2017multi}, and Multi-modal iSAM~\cite{hsiao2019mh}, they still cannot handle extremely ambiguous situations that cause state distributions with significant non-linearity and many modes.

\subsection{Non-parametric State Estimation for SLAM}

Monte Carlo estimation is a popular approach for localization and mapping in 2D environments due to its ability to represent multi-modal distributions, resulting in robust estimation. In particular, the particle filter~\cite{probrobo} represents the posterior distribution of an estimated state with a finite set of particles. While the particle filter has been widely used for sensor localization~\cite{probrobo}, it faces difficulties when applied to high-dimensional problems due to the curse of dimensionality. The RBPF SLAM~\cite{probrobo, grisetti2007improved} addressed this issue by removing the map from the state space and deriving a map estimate analytically from the sensor trajectory estimate. However, these filters suffer from the sample impoverishment problem~\cite{Arulampalam_2002}, caused by the resampling process that discards low-probability particles. This results in reduced estimation accuracy in highly ambiguous situations, where the state distribution is strongly multi-modal.

Recently, several works have introduced variational inference-based particle update strategies into the particle filter. The Stein particle filter~\cite{Maken_2022} leverages Stein variational gradient descent (SVGD)~\cite{NIPS2016_b3ba8f1b}, a variational inference algorithm that uses gradient information of a likelihood function to adjust samples, improving their fit to the likelihood function. In the Stein particle filter, particle states are updated based on SVGD, enhancing sampling efficiency. Koide et al. proposed a Gauss-Newton-based SVGD, which introduced Gauss-Newton-based state updates and GPU parallel processing into the Stein particle filter, successfully achieving rapid recovery from ``kidnapping'' situations by processing a large number of particles using a GPU~\cite{Koide_2024mega}. These methods have significantly improved sampling efficiency and demonstrated their applicability to high-dimensional problems. However, these variational inference-based particle filters have been applied only to localization and pose estimation problems. To our knowledge, our work is the first to apply the particle filter to a 6-DoF SLAM problem in a 3D environment, addressing the challenges of problem complexity and memory consumption.

\section{METHODOLOGY}
\subsection{Problem Setting and Algorithm Overview}

We estimate a time series of sensor poses ${\bm T}_t \in SE(3)$ and an environmental map $\mathcal{M}$ from point cloud scans $\mathcal{P}_t = \left\{ {\bm p}_k \in \mathbb{R}^3 \mid_{k = 1, \ldots, N^S} \right\}$ and IMU data $\mathcal{I}_t = \left[ {\bm a}_t, {\bm \omega}_t \right]$, where ${\bm a}_t \in \mathbb{R}^3$ and ${\bm \omega}_t \in \mathbb{R}^3$ are linear acceleration and angular velocity, respectively. As with a conventional particle filter, our system represents the state distribution with a set of samples (i.e., particles) $\mathcal{X}_t = \{ {\bm x}^1_t, \ldots, {\bm x}^N_t \}$.

{\bf Algorithm overview:} Similar to traditional RBPF SLAM, our system consists of three steps: prediction, correction, and representative state extraction. In the prediction step, we sample the current pose for each particle based on an estimate of relative sensor motion, which can be obtained using methods like LiDAR-IMU odometry. In the correction step, we update the estimates of the current pose and keyframe poses for each particle using gradient information from the likelihood function. Finally, we extract the representative state from the set of particles. To improve particle sampling efficiency and address the large memory footprint associated with map representation, we introduce keyframe-based map management and a gradient-guided particle state update strategy to the traditional RBPF SLAM. Each particle can be updated independently, and thus the entire update process can be efficiently performed on a GPU in parallel.

{\bf Particle state:} While the map estimate is analytically derived from a time series of sensor poses, naively storing an occupancy grid map for each particle would result in intractable memory consumption, as we handle a large number of particles and the memory footprint of occupancy grids increases substantially in 3D space. To reduce memory consumption, we introduce a keyframe-based map representation shown in Fig.~\ref{fig:system}. 
Keyframes are selected from past sensor frames at regular intervals, represented as $\{(\mathcal{P}_k, {\bm T}_{k}) |_{ k \in \mathcal{K}} \}$, where $\mathcal{K}$ are the indices of the keyframes. 
In our approach, each particle ${\bm x}_t^i := \{ {\bm T}^i_t \} \cup \{ {\bm T}^i_k |_{k \in \mathcal{K}} \}$ represents an estimate of the current sensor pose ${\bm T}_t^i$ and keyframe poses ${\bm T}_k^i$. The global map represented by particle ${\bm x}_t^i$ is given as a union of keyframes transformed in the world frame: $\mathcal{M}^i = \bigcup_{k \in \mathcal{K}} \text{transform}(\mathcal{P}_k, {\bm T}_k^i) $.

Because the point cloud data of the keyframes can be shared across all particles, the total memory footprint of the point cloud data increases only in proportion to the number of keyframes (i.e., independently of the number of particles). Although we still need to store the keyframe poses for each particle, these poses (4$\times$4 matrices) are lightweight, and the memory requirement remains manageable --- approximately 610 MB for 100,000 particles with 100 keyframes.

% We select some frames from a time series of the sensor frames as keyframe. The system retains point clouds corresponding to each keyframe. We syncronize the keyframe timestamps across all particles to let all particles use the same point cloud as ~\cite{fig:system}, which significantly decreases the required memory footprint of the map. Therefore, the state of particle is the current pose ${\bm x}_t^i:={\bm T}_t^i$ and the keyframe poses $\mathcal{K}_t^i = \left\{ {\bm x}_k^i := {\bm T}_k^i \mid_{k=1, \ldots, N^K} \right\}$. 

% In the proposed method, first, we sample the particles according to the motion model in the prediction step. Next, we evaluate the likelihood and its derivative between the neighbor keyframes and the current pose. If non-recent keyframes are included, the trajectory of each particle is corrected in the correction step. Then, we performe the importance weighting and keyframe adding. Finally, we remove unnecessary particles.

\subsection{Prediction Step}

In this step, we update the current sensor pose estimate for each particle based on relative motion prediction. To this end, we perform the loosely coupled LiDAR-IMU odometry implemented in~\cite{Koide_2024}, which combines scan-to-model GICP matching~\cite{segal2009generalized} and IMU preintegration~\cite{Forster_2015}. Given the estimated odometry trajectory ${\bm T}^o_t$, we obtain an estimate of the relative motion $\Delta{\bm T}_t \sim \left( {\bm T}^o_{t-1} \right)^{-1} {\bm T}^o_t$. We also estimate the covariance matrix of the motion model $\Sigma^{\Delta {\bm T}_t}$ by taking the inverse of the Hessian matrix of the GICP cost between consecutive frames. Then, the current pose estimate of particle ${\bm x}_t^i$ is updated as follows:
\begin{align}
    {\bm T}_t^i = {\bm T}_{t-1}^i \Delta {\bm T}_t \exp \left( {\bm \delta}_t^i \right),
\end{align}
where ${\bm \delta}^i_t \sim \mathcal{N} (0, \Sigma^{\Delta {\bm T}_t})$ is random noise in the tangent space of $\Delta {\bm T}_t$. 

\subsection{Correction Step}
\begin{figure}[t]
    \centering
    \includegraphics[width=0.8\linewidth]{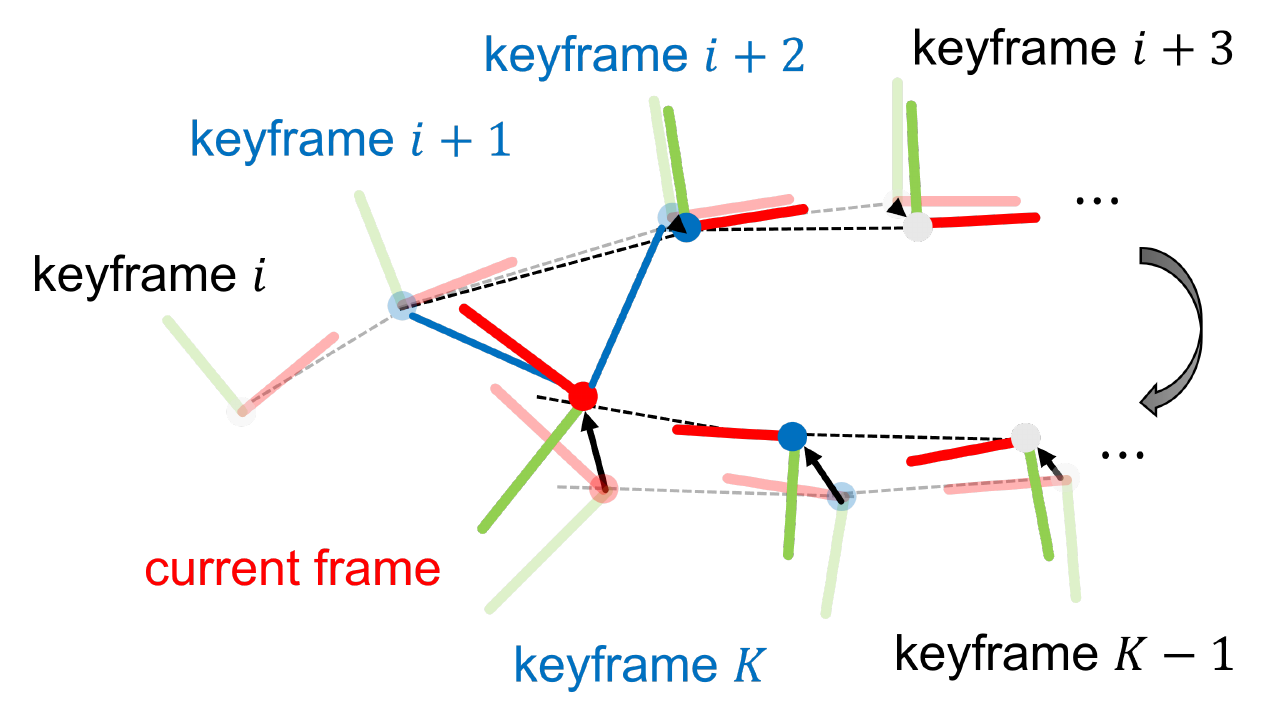}
    \caption{Relationship between the current frame and the neighbor keyframes, with three neighbor keyframes shown as an example. The likelihood is computed based on the registration error between the current frame and each neighbor keyframe. The current pose is then corrected using the gradient of the likelihood with respect to non-recent keyframes, specifically keyframes $i+1$ and $i+2$. The keyframes between the oldest neighbor keyframe and the newest keyframe are updated by propagating the current pose correction into them.}
    \label{fig:neighbor}
\end{figure}
%
% {\bf Likelihood Function}
\subsubsection{Likelihood Function}
To evaluate the consistency between the trajectory and the resultant map for each particle, we define the likelihood function based on the point cloud registration error. First, we select neighbor keyframes relative to the current frame based on the translation distance. We then compute the registration errors between the current frame and these neighbor keyframes. We use the distribution-to-distribution distance metric from GICP~\cite{segal2009generalized}, which models each point as a Gaussian distribution representing the local surface shape, with a voxel-based corresponding point search~\cite{Koide_2024}. The likelihood is defined as follows:
\begin{align}
    &\log p(\mathcal{P}_t | {\bm x}_t^i) = \sum_{k \in \mathcal{K}_t^i} \log p(\mathcal{P}_t, \mathcal{P}_k| {\bm T}_t^i, {\bm T}_k^i), \\ 
    &\log p(\mathcal{P}_t, \mathcal{P}_k | {\bm T}_t^i, {\bm T}_k^i) = - \sum_{{\bm p}_j \in \mathcal{P}_t} {\bm e}_{j}^\top {\bm \Omega}_{j} {\bm e}_{j}, \\
    &{\bm e}_{j} = {\bm \mu}_j^{\prime} - ^k{\bm T}_t^i {\bm \mu}_j, \quad {\bm \Omega}_j = \left( {\bm \Sigma}_j^{\prime} +  ^k{\bm T}_t^i {\bm \Sigma}_j ( ^k{\bm T}_t^i)^\top \right)^{-1},
\end{align}
where $\mathcal{K}_t^i$ is the index set of neighbor keyframes, ${\bm p}_j \sim \mathcal{N}({\bm \mu}_j, {\bm \Sigma}_j)$ is a point of the current point cloud $\mathcal{P}_t$, ${\bm p}_j^{\prime} \sim \mathcal{N}({\bm \mu}_j^{\prime}, {\bm \Sigma}_j^{\prime})$ is the corresponding point with ${\bm p}_j$ of a neighbor keyframe point cloud $\mathcal{P}_k$, and $^k{\bm T}_t^i = \left( {\bm T}_{k}^i \right)^{-1} {\bm T}_{t}^i$ is the relative pose between ${\bm T}_k^i$ and ${\bm T}_t^i$.

% {\bf Particle State Update}
\subsubsection{Particle State Update}
For each particle, if an old keyframe is found among the three neighbor keyframes (i.e., a loop is detected), we update the state of that particle, including the current sensor pose and past keyframe poses, to ensure a better fit to the likelihood function. To achieve this, we implement a two-step state update scheme. First, we update the current pose estimate to maximize the log likelihood. Then, we correct the keyframe poses to maintain the consistency of the overall trajectory.

{\it Current pose update: }
To update the current pose, we compute the optimal update vector ${\bm \psi}^i$ to maximize the log likelihood based on Gauss-Newton optimization as follows:
\begin{align}
    {\bm \psi}^i &= \left( {\bm H} \right)^{-1} {\bm b},
\end{align}
\begin{align}
{\bm H} &= \sum_j {\bm J}_j^\top {\bm \Omega}_j {\bm J}_j, &{\bm b} &= \sum_j {\bm J}_j^\top {\bm \Omega}_j {\bm e}_j, & {\bm J_j} &= \frac{\partial {\bm e}_j} {\partial ^k{\bm T}_t^i}.
\end{align}
We then update the current pose estimate using ${\bm \psi}^i$ by applying this update vector:
\begin{align}
    {{\bm T}_{t}^i}' &= {\bm T}_t^i  \exp \left( {\bm \psi}^i \right).
\end{align}

{\it Keyframe pose update: }
To correct estimation drift, we propagate the update of the current pose estimate ${\bm \psi}^i$ to past keyframe poses. Because we process a large number of particles on a GPU and cannot perform rigorous trajectory optimization for each particle, we implement the simplified loop closure method shown in Fig.~\ref{fig:neighbor}. This method applies the update vector to keyframe poses, with the weights discounted according to the travel distance as follows:
\begin{align}
    {\bm \psi}_k^i &= \frac
    { d\left( {\bm t}_{t_k}, {\bm t}_{t_o} \right) }
    { d\left( {\bm t}_t, {\bm t}_{t_o} \right) }
    {\bm \psi}^i, \\
    d\left( {\bm t}_s, {\bm t}_o \right) &= \sum_{t=o}^s \|\ {\bm t}_{t+1}^i - {\bm t}_t^i \|,
\end{align}
where $t_o$ is the time step of the oldest neighbor keyframe among the selected ones, $t_k$ is the time step of keyframe $k$, and ${\bm t}_t^i$ is the translation of ${\bm T}_t^i$. Path length $d$ is defined as the travel distance between two time steps estimated by LiDAR-IMU odometry. The partial sum of the path length, divided by the total path length, is multiplied by ${\bm \psi^i}$ to obtain the update displacement vector of each keyframe. We then refine the estimate for each keyframe's pose as follows:
\begin{align}
    {{\bm T}_{k}^i}' &= {\bm T}_k^i  \exp \left( {\bm \psi}_k^i \right).
\end{align}

\subsection{Importance Weighting}
%過去の尤度を伝播させていることを
To extract the representative particle, we compute the weight for each particle according to the importance sampling principle~\cite{grisetti2007improved}. The consistency between the current pose and the map is expressed by the likelihood. Considering not only the current pose but the entire trajectory, we compute the product of the likelihoods as follows:
\begin{align}
    p ({\bm x}_{1:t}^i | \mathcal{P}_{1:t}) = p (\mathcal{P}_t | {\bm x}_t^i) p ({\bm x}_{1:t-1}^i | \mathcal{P}_{1:t-1}).
\end{align}
%

% Second, to consider the trajectory correction, we compute the relative pose error between each keyframe for odometry.
% %
% \begin{align}
%     \log p_m \left({\bm x}_{1:t}^i | \mathcal{P}_{1:t} \right) =  \sum_k \log \left( \left( {\bm T}_k^i \right)^{-1} {\bm T}_{k+1}^i \Delta {\bm T}_k \right),
% \end{align}
% %
% where $\Delta{\bm T}_t \sim \left( {\bm T}^o_{t-1} \right)^{-1} {\bm T}^o_{k+1}$

\subsection{Updating Keyframe List}

To avoid overly frequent insertion of keyframes and to maintain a moderate number of keyframes, we use the overlap rate between the current frame and the last keyframe. We transform the current point cloud into the coordinate frame of the last keyframe using the relative motion estimated by LiDAR-IMU odometry and evaluate the overlap rate with an efficient voxel-based overlap metric~\cite{Koide_2024}. If the overlap rate falls below a specified threshold (e.g., 70\%), the current frame is inserted into the keyframe list.

\begin{figure}[tb]
 \centering
 \includegraphics[width=0.7\linewidth]{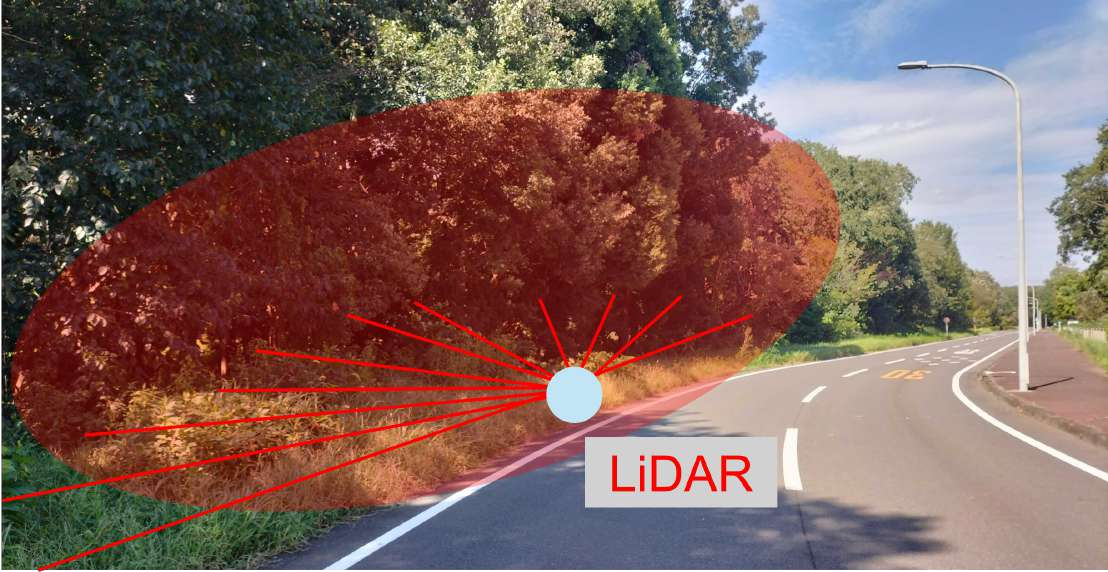}
 \caption{Forest-like environment for the outdoor experiment. The loop was detected while the LiDAR was facing the rows of trees. The repeated and complex geometries cause multiple uncertain loop candidates and make it difficult to robustly identify correct loops. We cropped the points in the backward direction in the LiDAR frame to imitate a challenging mapping situation that limits the field of view of the LiDAR.}
 \label{fig:forest}
\end{figure}

\begin{table}[tb]

    \centering
    \caption{Absolute trajectory errors for forest-like sequences}
    \resizebox{80mm}{!}{%
    \begin{tabular}{c|cccc}
        \toprule
    Metric      & SLICT          & DLIOM           & GLIM            & Proposed \\
    \midrule
    ATE~[m]      & 11.2 $\pm$ 4.68  & 3.44 $\pm$ 1.90   & 5.67 $\pm$ 2.04   & \bf{2.58 $\pm$ 1.03} \\
        \bottomrule
    \end{tabular}%
    }
    \label{tab:results}
\end{table}

\subsection{Dead Particle Pruning}

Although the gradient-guided particle update significantly improves sampling efficiency, some particles may still fall into unlikely states due to the non-linearity and non-convexity of the likelihood function. Because the weights of these ``dead'' particles eventually become zero, rendering them useless for state estimation, we prune them and spawn new particles using the weights of the remaining particles. While this pruning strategy may appear similar to traditional resampling, it serves a different purpose to remove only completely unusable particles (e.g., ones with a log likelihood smaller than $1 \times 10^{-16}$ or a posterior smaller than $1 \times 10^{-8}$).

\begin{figure}[tb]
 \centering
 \includegraphics[width=0.75\linewidth]{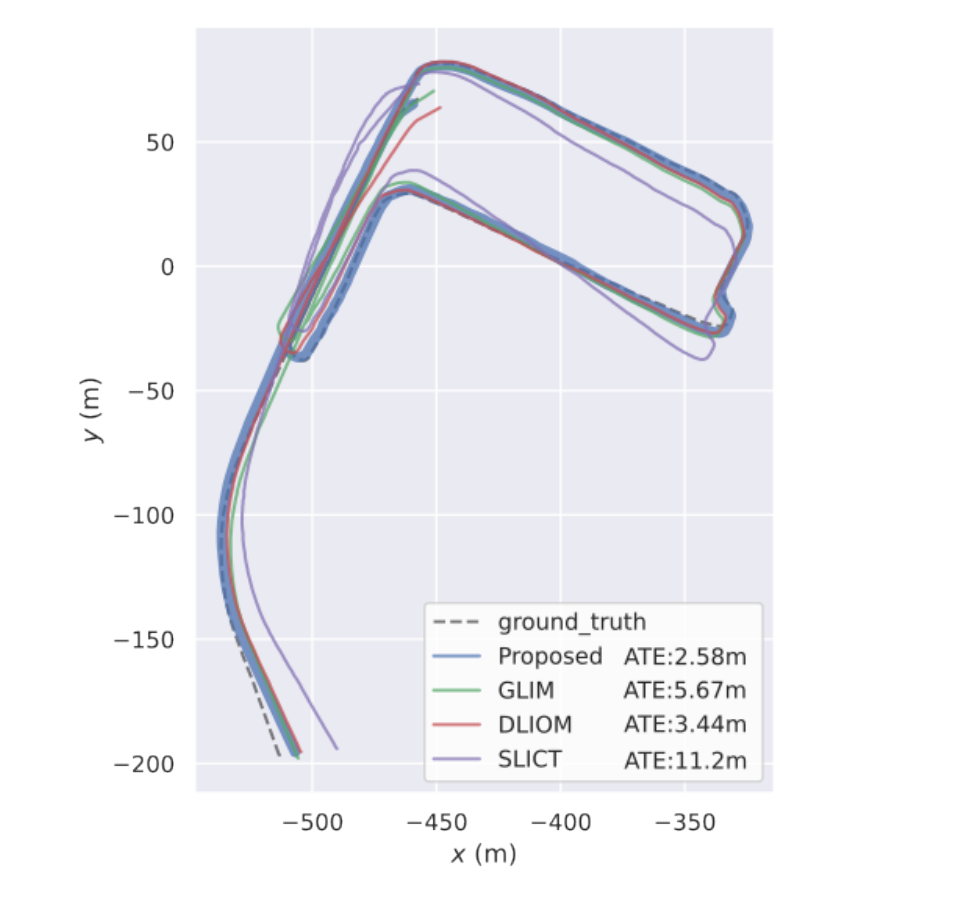}
 \caption{Estimated trajectories for the forest-like experiment.}
 \label{fig:evo}
\end{figure}

\section{EXPERIMENT}

\subsection{Loop Closure in a Forest-like Environment}
{\bf Experimental setting:} To demonstrate the ambiguity resolution ability of the proposed method in highly ambiguous situations, we conducted an experiment in the environment shown in Fig.~\ref{fig:forest}. In the forest-like environment, there are repeated rows of trees causing multiple uncertain loop candidates. We used a Livox MID-360 to acquire point clouds and IMU measurements. To imitate the difficulty of loop closing in forest-like environments, we cropped the points in the backward direction in the LiDAR frame so that it could observe only points in the forest-like region.
In the proposed method, 100,000 particles were processed in real-time using an NVIDIA GeForce RTX 4090. As a baseline, we also tested three graph-based SLAM methods: SLICT, which is based on continuous-time sliding window odometry estimation and loop closure with pose graph optimization~\cite{Nguyen_2023}; DLIOM, which utilizes direct scan matching-based odometry and submap registration-based trajectory optimization~\cite{wang2022d}; and GLIM, which is based on GPU-accelerated sliding window odometry and global matching cost minimization for trajectory optimization~\cite{Koide_2024}.

% To deal with the estimation drift in the Z direction, we added an altitude error between the current sensor pose and neighbor keyframe poses to the likelihood function of the proposed method. We also added a similar constraint to the factor graph of GLIM \cite{Koide_2024}.

To obtain a smooth and drift-free reference trajectory as ground truth, we manually aligned the scan point clouds with a map point cloud and conducted batch optimization of the point cloud registration errors and IMU motion errors. We used precise survey map data created by a mobile mapping system\footnote{The reference map data are available in AIST 3DDB: \url{https://gsrt.digiarc.aist.go.jp/3ddb_demo/tdv/index.html}.}. The estimated trajectories were evaluated with the absolute trajectory error (ATE) metric \cite{Zhang_2018} using the evo toolkit\footnote{\url{https://github.com/MichaelGrupp/evo}}. For the proposed method, we selected the particle with the largest weight as the representative particle and evaluated its ATE.

{\bf Experimental result:} Table~\ref{tab:results} summarizes the ATEs for the evaluated methods. The existing methods based on factor graph optimization struggled in environments with rows of trees and exhibited large ATEs  (SLICT: 11.2 $\pm$ 4.68 m, DLIOM: 3.44 $\pm$ 1.90 m, GLIM: 5.67 $\pm$ 2.04 m). Note that all existing methods detected incorrect loops due to the repeated and similar geometric features in the environment, leading to degraded trajectory correction. The proposed method achieved the best ATE of $2.87 \pm 1.03$ m. It gradually spread the particles based on a random walk, using covariance matrices provided by the odometry estimation until the loop was closed. Once closed, the trajectory for each particle was corrected, and the particle with the most likely match was successfully selected as the representative particle. Although incorrect trajectories were occasionally chosen as the representative particle due to high ambiguity, the correct representative particle was eventually selected when sufficient observations were made. Fig.~\ref{fig:evo} displays the estimated trajectories, where the trajectories of the existing methods show large errors due to incorrect loop closing, while the representative particle of the proposed method demonstrates smaller errors.

\begin{figure}[tb]
 \centering
 \includegraphics[width=0.9\linewidth]{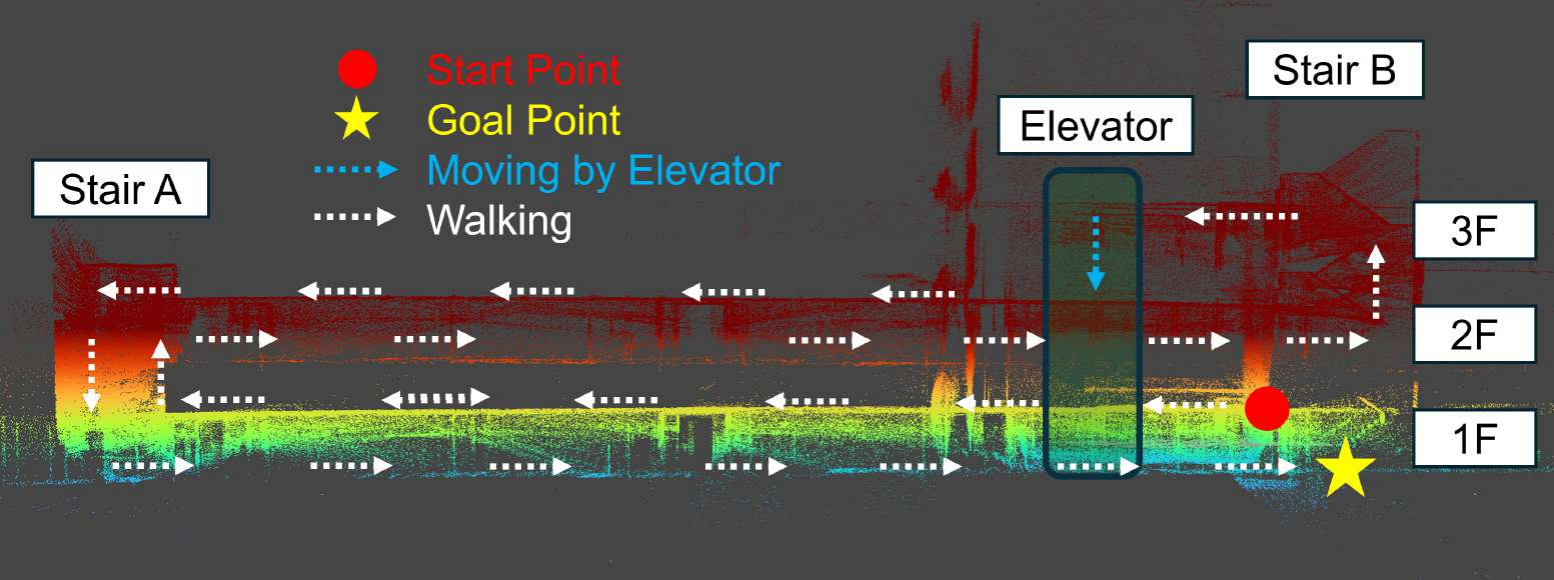}
 \caption{Point cloud map successfully created by the proposed method during the elevator experiment. The sensor's path in this experiment is shown on the X-Z plane. Starting from the red dot, indicating the starting point, the sensor moved to the third floor through corridors and on stairs. It then descended to the second floor using the elevator. While it was inside the confined elevator, LiDAR-IMU odometry estimation was no longer able to provide valid motion estimates. Finally, the sensor moved from the second floor to the first floor via the stairs.}
 \label{fig:elevator}
\end{figure}

\subsection{Elevator Experiment}

\begin{figure}[tb]
  \centering
  \includegraphics[width=0.9\linewidth]{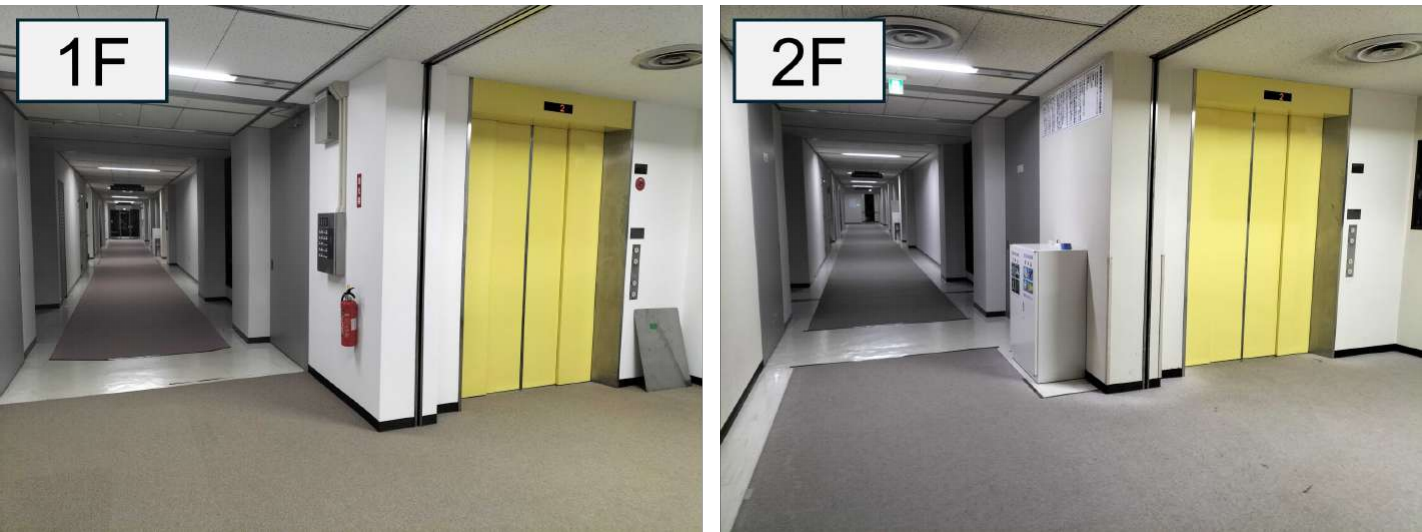}
  \caption{Snapshots of the elevator environment. Each floor has a very similar floor plan, posing an extreme challenge for loop closures after moving between floors using the elevator.}
  \label{fig:floors}
\end{figure}

\begin{figure}[tb]
 \centering
 \includegraphics[width=0.9\linewidth]{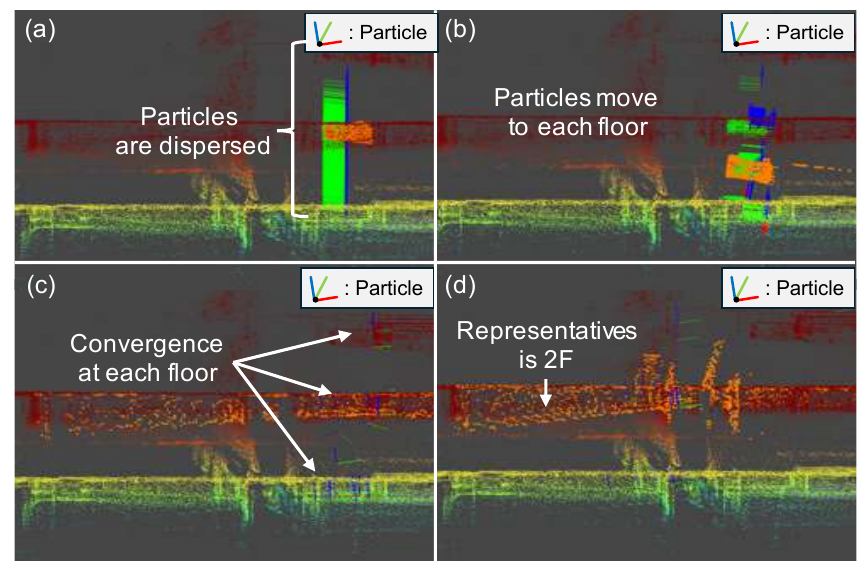}
  \caption{Transition of the particle set in the elevator experiment. While moving in the elevator, the particles were vertically distributed according to a heuristic elevator transition model, as shown in (a). After exiting the elevator, the particle poses were adjusted to maintain consistency with past keyframes, as shown in (b). The particles then split into three hypothesis groups, as shown in (c) and (d).}
  \label{fig:particles}
\end{figure}

{\bf Experimental setting:} To demonstrate the ability of the proposed method to close loops during a severe kidnapping situation, we conducted an experiment in an indoor, multi-floor environment with an elevator. All floors had very similar floorplans, as shown in Fig.~\ref{fig:floors}. We recorded a sequence in which the sensor moved from the first floor to the third floor through corridors and stairs, and then moved to the second floor using the elevator, as shown in Fig.~\ref{fig:elevator}.

While the sensor was moving to the second floor via the elevator, we employed a simple heuristic method to detect the elevator situation based on the threshold of the median of point distances, spreading the particles using a random walk in the vertical direction. While this minimal heuristic was designed specifically for elevator scenarios, a more sophisticated pose prediction method (e.g., learned IMU odometry~\cite{liu2020tlio}) could be applied to a wider range of kidnapping situations.

{\bf Experimental result:} Fig.~\ref{fig:elevator} shows the mapping result of the proposed method. It successfully closed the loop after the severe kidnapping situation caused by the use of the elevator. While in the elevator, the particles were dispersed vertically, as shown in Fig.~\ref{fig:particles}~(a). After the sensor exited the elevator, the particles moved toward high-likelihood locations (Fig.~\ref{fig:particles}~(b)). In Figs.\ref{fig:particles}~(c) and (d), three major hypothesis groups formed across different floors, demonstrating the proposed method’s ability to represent extremely multi-modal distributions with high ambiguity. Ultimately, the particle with the correct trajectory hypothesis was selected as the representative particle, as shown in Fig.~\ref{fig:elevator}. Note that this experiment posed a challenge far beyond the estimation capabilities of existing methods due to the severity of the kidnapping situation.

{\bf Processing time:} The average processing times for the outdoor and indoor experiments were approximately 60 ms and 50 ms per frame, respectively, on an NVIDIA GeForce RTX 4090. These times were well within real-time requirements (100 ms per frame).

\section{CONCLUSIONS}

This paper presented 6-DoF range-based Monte Carlo SLAM. The combination of a keyframe-based map representation, a gradient-guided particle update strategy, and GPU computation significantly improved sampling efficiency and enhanced the ability to represent state ambiguity. These advancements enabled efficient and robust mapping, even in extreme kidnapping situations.

%%%%%%%%%%%%%%%%%%%%%%%%%%%%%%%%%%%%%%%%%%%%%%%%%%%%%%%%%%%%%%%%%%%%%%%%%%%%%%%%

\balance

%%%%%%%%%%%%%%%%%%%%%%%%%%%%%%%%%%%%%%%%%%%%%%%%%%%%%%%
\bibliographystyle{IEEEtran}

\end{document}